\documentclass[10pt,twocolumn,letterpaper]{article}

\usepackage{wacv}              

\definecolor{wacvblue}{rgb}{0.21,0.49,0.74}
\usepackage[pagebackref,breaklinks,colorlinks,allcolors=wacvblue]{hyperref}
\usepackage{multirow}
\usepackage{comment}
\usepackage[accsupp]{axessibility}  

\def\wacvPaperID{*****} 
\def\confName{WACV}
\def\confYear{2026}

\title{FujiView: Multimodal Late-Fusion for Predicting Scenic Visibility}

\author{Bryceton Bible$^{1}$ \qquad Shah Md Nehal Hasnaeen$^{1}$ \qquad Hairong Qi$^{1}$ \\
$^{1}$University of Tennessee, Knoxville \\
{\tt\small bbible@vols.utk.edu, shasnaee@vols.utk.edu, hqi@utk.edu}
}

\begin{document}
\maketitle
\begin{abstract}
Visibility of natural landmarks such as Mount Fuji is a defining factor in both tourism planning and visitor experience, yet it remains difficult to predict due to rapidly changing atmospheric conditions. We present \textbf{FujiView}, a multimodal learning framework and dataset for predicting scenic visibility by fusing webcam imagery with structured meteorological data. Our late-fusion approach combines image-derived class probabilities with numerical weather features to classify visibility into five 
categories. The dataset currently comprises over 100,000 webcam images paired with concurrent and forecasted weather conditions from more than 40 cameras around Mount Fuji, and continues to expand; it will be released to support further research in environmental forecasting. Experiments show that YOLO-based vision features dominate short-term horizons such as ``nowcasting''  and ``samedaycasting'', while weather-driven forecasts increasingly take over as the primary predictive signal beyond $+1$d. Late fusion consistently yields the highest overall accuracy, achieving $\mathrm{ACC} \approx 0.89$ for same-day prediction and up to \textbf{84\%} for next-day forecasts. These results position \textbf{Scenic Visibility Forecasting (SVF)} as a new benchmark task for multimodal learning.
\end{abstract}
    
\section{Introduction}
\label{sec:intro}

\subsection{Motivation}

\begin{figure}
    \centering
    \includegraphics[width=0.75\linewidth]{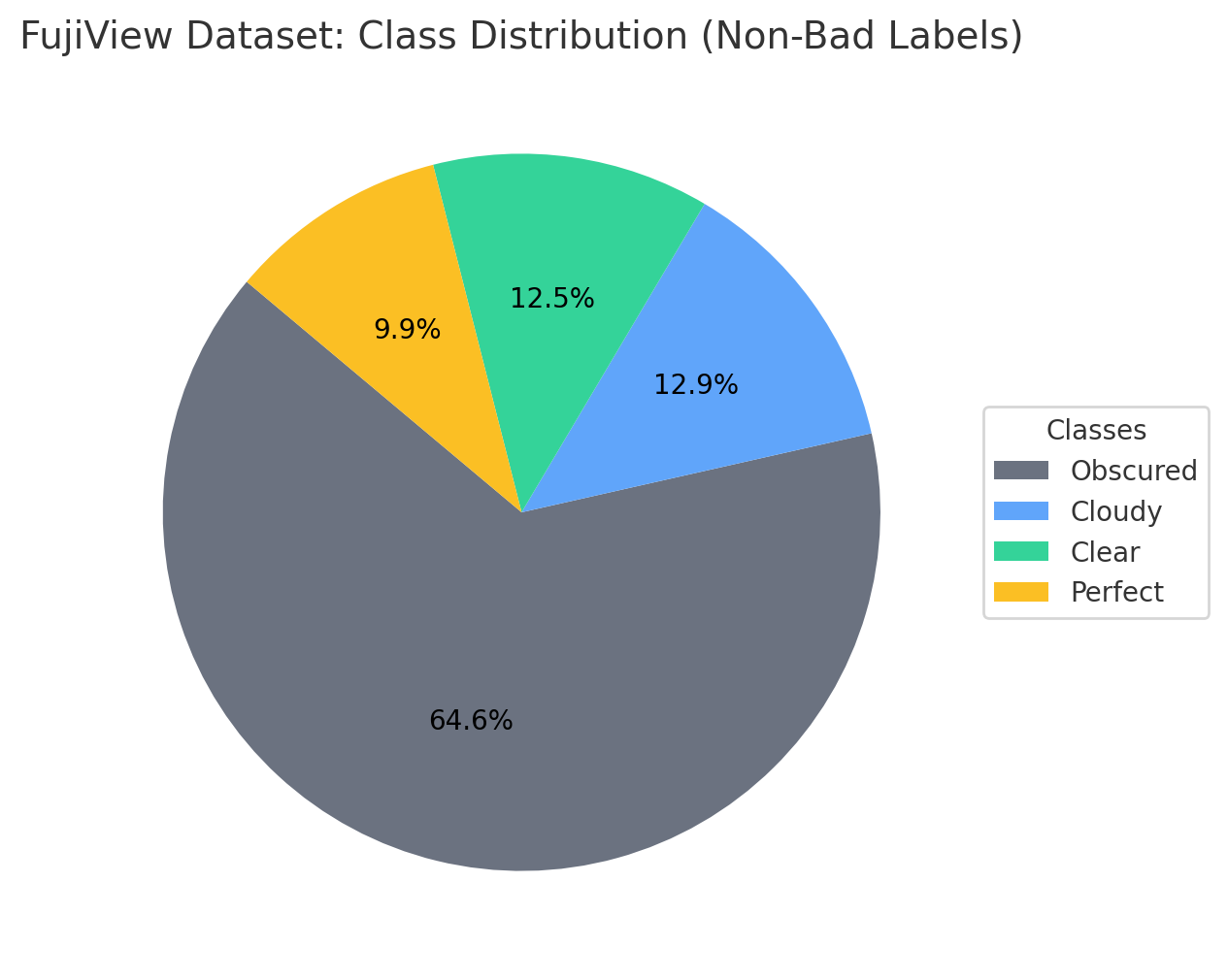}
    \caption{Measured class distribution for Fuji visibility}
    \label{fig:class-dist-pie}
\end{figure}

Mount Fuji is one of Japan’s most iconic and sought-after natural landmarks. Due to Japan’s efficient and affordable rail network, tourists can easily reach multiple observation points—including Fuji City, Shizuoka, or the Fuji Five Lakes region near Kawaguchiko—from urban centers like Tokyo in under two hours. However, Mount Fuji’s visibility is notoriously unpredictable, especially during the rainy season (tsuyu) or summer months when cloud cover is frequent. While Mount Fuji is at least partially visible from long distance up to 36\% of days per year \cite{pubmedtokyo}, up close good visibility is rare. Our dataset's measured class distribution (See Figure \ref{fig:class-dist-pie}) lines up with these long distance visibility statistics, as well as anecdotal claims that approximately 20\% of days per year offer ``good enough'' visibility of the mountain.  This presents a major challenge to both casual travelers and photography enthusiasts hoping to glimpse or capture the mountain during their limited time in the country.

In such scenarios, a traveler may wake up with multiple viable destinations but little clarity on where Mount Fuji might be visible. They may be forced to choose between acting immediately with limited visibility or waiting in hopes of better conditions later—potentially missing their only chance. This uncertainty not only affects individual experiences but may also result in missed opportunities for local tourism. By providing more reliable, data-driven visibility information, our system has the potential to guide visitors toward regions with favorable conditions, thereby distributing tourism more effectively across lesser-known but equally accessible towns surrounding Mount Fuji.

Such an approach is especially timely given the rise \cite{jnto2023tourism,jnto2024record} in inbound tourism to Japan during peak seasons \cite{tonomura2015seasonality}—many of which coincide with low visibility periods—and the country’s broader push for regional revitalization. By encouraging travel to areas that may otherwise be overlooked, particularly on days when weather conditions are optimal, this system can support both visitor satisfaction and economic development in peripheral communities.

This project seeks not only to provide real-time visibility assessment using webcam feeds, but also to forecast future visibility conditions by combining visual observations with current and predicted meteorological data. Using a multimodal late-fusion approach, our system learns visibility trends across time, enabling informed decisions about when and where Mount Fuji will be visible. The distinction between \textit{nowcasting} (image-only assessment of present conditions) and true \textit{forecasting} (predicting visibility hours or days ahead) is central to our design, and highlights the gap our work fills between existing methods and practical traveler needs.

Beyond Mount Fuji, the techniques developed in this work could be extended to other natural landmarks and scenic destinations where visibility is critical. The project thus serves both practical and scientific purposes: improving the travel experience through intelligent guidance, supporting seasonal tourism and regional revitalization, and contributing to research on multimodal learning, real-time perception, and environmental forecasting.

\subsection{Challenges}
While Mount Fuji is visible from many vantage points across central Japan, yet its appearance is highly sensitive to local weather and cloud formation. Although webcam feeds and weather forecasts are widely available, no system currently integrates these sources into actionable visibility predictions. Existing tools provide either raw webcam footage or generic forecasts, but do not answer the concrete question of whether the mountain will be visible at a given place and time.

Visibility is also not a binary condition but a continuum—ranging from full clarity to partial occlusion, haze, or complete cloud cover. Predicting this spectrum from image and weather data poses challenges such as temporal variation, atmospheric noise, and the fusion of heterogeneous modalities. Moreover, the absence of labeled datasets and standard benchmarks for scenic visibility further complicates model development and evaluation.

\begin{figure}[ht]
    \centering
    \includegraphics[width=0.9\linewidth]{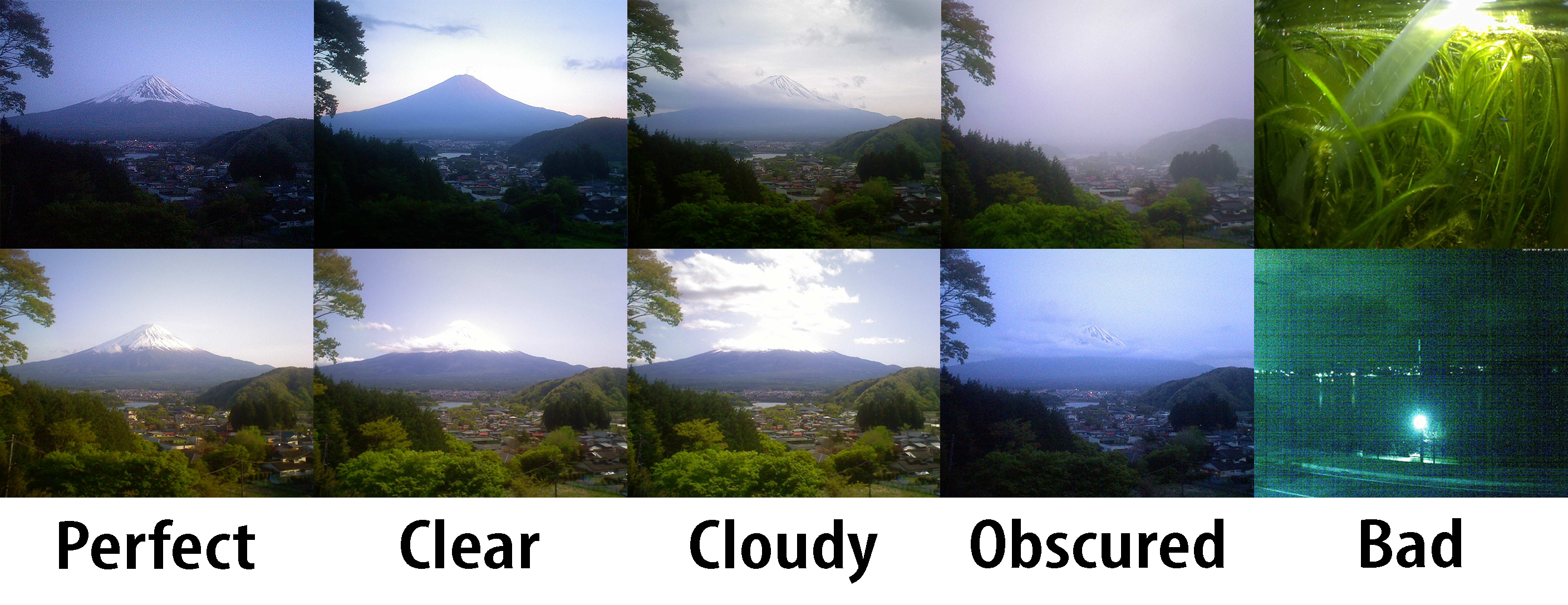}
    \caption{Mount Fuji visibility classes}
    \label{fig:classes}
\end{figure}

This project addresses the gap by building a multimodal learning system capable of performing both real-time nowcasting and short-term forecasting of visibility using fused webcam and meteorological data. We formalize this problem as \textbf{Scenic Visibility Forecasting (SVF)}: predicting the perceptual visibility of natural landmarks from multimodal inputs across varying horizons (e.g., immediate nowcasts, same-day samedaycasts, or next-day tomorrowcasts). Unlike generic weather prediction, SVF directly targets \textit{visibility as experienced by humans}, with outputs defined along a preceptual spectrum (perfect, clear, cloudly, obscured, as shown in Figure~\ref{fig:classes}) or collapsed into binary ``visible'' vs. ``not visible'' categories. Framing SVF as a benchmark task highlights both the practical importance for travelers and the scientific challenges for multimodal learning.

In short, FujiView is designed to help answer the practical question: ``\textbf{Can I see Mt. Fuji today? Where? What about tomorrow?}'' At the same time, 
SVF
 is intended to generalize this question beyond Fuji to any natural landmark where visibility matters, and to provide a benchmark for evaluating multimodal techniques in this domain.

\subsection{Contributions}

This work makes the following contributions:

\begin{itemize}
    \item \textbf{Large-scale and growing multimodal dataset.} We are constructing a continually expanding dataset of webcam images paired with concurrent and forecasted weather conditions from more than 40 locations around Mount Fuji. As of this writing, over 113,000 images have been collected and approximately 26,000 have been manually labeled, with the dataset projected to exceed 320,000 images by the year's end. This resource will be released publicly and updated over time to support the development of SVF as a benchmark task.
    \item \textbf{Fusion-based modeling framework.} We develop late-fusion models that integrate pretrained image classifiers (e.g., YOLOv8) with structured meteorological features, showing complementary strengths across nowcasting and forecasting. 
Our experiments reveal that image features dominate same-day prediction, achieving the highest accuracy of $0.89$ at $+0$d, but their relative contribution decreases with longer horizons as weather forecasts become more predictive. 
Late fusion yields accuracies of $0.84$ at $+1$d, $0.77$ at $+2$d, and $0.72$ at $+3$d, demonstrating its advantage for multi-day forecasts.

    \item \textbf{Scenic Visibility Forecasting (SVF) as a benchmark task.} We formally define the prediction of landmark visibility from multimodal inputs as a new challenge for the vision community, emphasizing the distinction between nowcasting and forecasting.
    \item \textbf{Tools and deployment.} We provide a web-based data exploration and labeling platform, automated pipelines for large-scale multimodal collection, and a public-facing web application demonstrating real-world impact.
\end{itemize}

To our knowledge, this is the first work to systematically study multimodal forecasting of landmark visibility, and the first to introduce a dataset at this scale for this task.

\section{Related Works}

The application of computer vision to weather-related tasks is not entirely new. One of the earliest and most common use cases pertains to road vehicles, such as the work by Ahmed \textit{et al.}~\cite{ahmed2023weather}. After collecting images from roadside and car dash cameras, they used a pretrained convolutional neural network (CNN) to classify the weather conditions in each image as dry, rainy, snowy, foggy, and more. While their CNN was accurate and consistent, it only focused on current weather conditions and could not forecast future troubling weather conditions.

Some research groups simplified the problem from identifying specific weather conditions to focusing on meteorological visibility, or the maximum distance from which objects can be seen clearly. One example of such work belongs to Li, Fu, and Lo~\cite{li2017_visibility}. They collected images from Tsim Sha Tsui, Hong Kong, and attempted to predict visibility. They used CNNs for feature extraction and then a general regression neural network (GRNN) to generate visibility predictions. Ultimately, they achieved about 62\% accuracy on their predictions, outperforming traditional methods. However, they noted that accuracy might be improved with better quality images and more environmental context.

Reyes also focused on predicting visibility using CNNs~\cite{reyes2023_visibility},  but emphasized creating an approach that would be generalizable to images from new cameras in new locations.  However, the approach struggled  with label noise in the dataset and it was suggested that hand labeling images might be a potential solution.

In recent years, computer vision techniques have been applied to more than identifying visibility and type of weather. Ongkittikul \textit{et al.}~\cite{ongkittikul2024forecasting} attempted to improve weather forecasting by analyzing images of clouds. Specifically, they used the Edge Flow algorithm to convert the cloud images into vector data to train a YOLO model. This helped overcome the tendency of sky images to be skewed toward certain classes. Ultimately, this model classified clouds into one of five categories with a mean average precision (mAP) of almost 85\%, enabling forecasts to be made.

Paletta \textit{et al.}~\cite{paletta2023solar} explored a different type of forecasting with computer vision. They passed cloud images into CNNs and long short-term memory (LSTM) networks to generate solar forecasts. In other words, they focused on predicting cloud movement and whether the sun would be blocked, and they did experience success.

With our work, we aim to use webcam images and weather forecasts to predict the future visibility of Mount Fuji. Some of the issues experienced in previous visibility research included poor image quality, lack of environmental context, and struggles with dataset creation. To overcome image quality, we are using images from over 30 cameras around Mount Fuji. To ensure we do not lack environmental context, we are not just focusing on images—we also incorporate past, present, and forecasted weather data. Finally, to avoid mislabeled data, we manually label each image, a time-consuming but high-quality approach.

\section{Methodology}
\subsection{Dataset}

\begin{figure*}[!ht]
    \centering
    \includegraphics[width=0.75\linewidth]{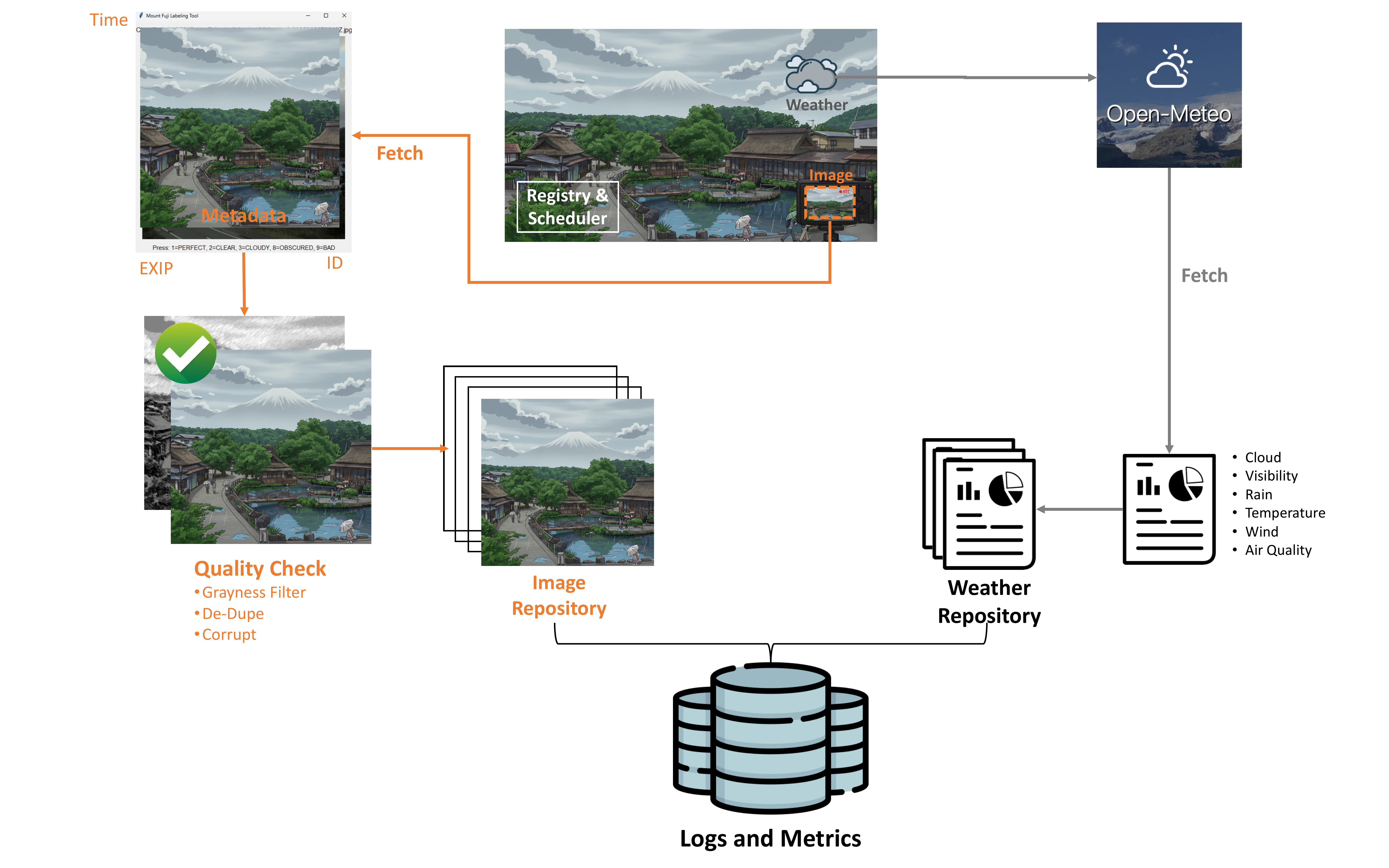}
    \caption{Data collection pipeline}
    \label{fig:data-pipeline}
\end{figure*}
For visibility forecasting of Mount Fuji, we needed both image and weather data. To obtain the images, we utilized 42 live webcams from various regions around Mount Fuji, saving an image of the feed from each webcam every 30 minutes. We then manually assigned one of five classes to each image:
\begin{itemize}
    \item \textbf{Perfect} - Mount Fuji fully visible with no cloud overlap.
    \item \textbf{Clear} - Visible with minimal cloud overlap.
    \item \textbf{Cloudy} - Visible but with substantial cloud overlap.
    \item \textbf{Obscured} - Mostly or completely hidden by clouds or atmospheric effects (unlikely to be worth the trip).
    \item \textbf{Bad} - Image unusable due to darkness, camera malfunction, or other failures.
\end{itemize}

Weather data was collected via Open-Meteo's APIs at the same 30-minute intervals. For each timestamp we recorded current conditions, same-day forecasts, previous-day conditions, and forecasts for one to six days ahead (experiments in this paper use +0-3d), aligned to corresponding webcam images, such that each dataset entry includes both image-level metadata and aligned weather variables.

\textbf{Image Variables:}  Assigned visibility label, model predictions from YOLOv8 (at late-fusion stage), image metadata (e.g., webcam identifier, lat/lon, capture time)

\textbf{Weather Variables} (from Open-Meteo, including but not limited to): temperature, categorical weather code, humidity, precipitation (snow/rain), cloud cover, surface and sea level pressure, wind speed/direction, etc.

\subsection{Data Collection and Annotation Pipeline}
The data collection pipeline (Figure~\ref{fig:data-pipeline}) integrates visual and meteorological streams into a synchronized multimodal dataset. A central scheduler triggers workers every 30 minutes to fetch new webcam imagery and weather data.
\textbf{Images} are collected from 42 publicly available webcams via a scraping and video extraction layer. Metadata (timestamp, camera ID, EXIF fields) are attached, followed by automated quality control checks (de-duplicating, image corruption/emptiness check). Images and metadata are stored in a raw image repository with logs tracking uptime and latency.
\textbf{Weather} is concurrently retrieved via the Open-Meteo API, and includes current and forecasted conditions (including temperature, humidity, precipitation, cloud cover, wind and pressure). These features are stored in a weather repository aligned by timestamp with the image data.

For ground-truth labeling, we developed a Python-based annotation tool to rapidly classify each image into \texttt{PERFECT}, \texttt{CLEAR}, \texttt{CLOUDY}, \texttt{OBSCURED}, and \texttt{BAD}. As part of quality control, images exceeding a 40\% grayness threshold were automatically flagged as \texttt{BAD} to filter out nighttime or otherwise unusable frames, with human review to avoid false rejections. We then fine-tuned a YOLOv8 classifier on this curated dataset and use its calibrated softmax probabilities $P(\texttt{PERFECT})$, $P(\texttt{CLEAR})$, $P(\texttt{CLOUDY})$, and $P(\texttt{OBSCURED})$ as continuous vision features rather than hard class predictions, allowing downstream models to leverage uncertainty in ambiguous cases.

To prepare training targets, we aggregate per-frame predictions into day-level binary labels indicating whether at least $\theta=50\%$ of frames for that (camera, date) were classified as visible (\texttt{CLEAR} or \texttt{PERFECT}). These day-level labels are shifted forward to create multi-horizon targets ($+0$d through $+3$d). To reduce label leakage for same-day forecasting, we construct either a \textit{first-frame} snapshot or a \textit{3-hour morning window} of features per day, allowing us to evaluate whether using a small time window improves predictive performance. The resulting fused dataset combines YOLO probabilities, aggregated labels, and aligned meteorological features (current and forecast) and is used to train the late fusion models described in Section~\ref{sec:lightgbm-arch}.

\subsection{The Late Fusion Model}
\label{sec:lightgbm-arch}
\paragraph{Motivation}
While webcam imagery alone is capable of providing instantaneous visibility classification \citep{li2017_visibility,reyes2023_visibility}, visibility forecasting (even if only \textit{samedaycasting} visibility from the current time onward) relies on meteorological variables that encode future trends in conditions like cloud cover, precipitation, and pressure \citep{kim2022_shortterm_vis,chen2024_gcn_gru}. These variables are not directly observable from static images, and conversely weather forecasts cannot capture fine-grained occlusion patterns visible in real-time imagery. A late-fusion strategy allows the model to leverage both modalities in the roles they are most suited for—vision for immediate state assessment, and meteorology for forward prediction—without forcing them into a joint feature space that may obscure these complementary strengths. In contrast, early fusion of raw pixel-level features with weather variables risks over-parameterization and higher sample complexity \citep{baltrusaitis2019_mmsurvey,snoek2005_earlylate}, and typically requires more data to reliably generalize. By combining probability outputs (YOLO softmax probabilities) with structured, aligned weather variables via late fusion, we obtain a tractable, interpretable mechanism that matches our dataset size and task design.

LightGBM was selected as the fusion learner because as a gradient-boosted decision tree model, it is optimized for efficiency on heterogeneous tabular data. Unlike MLP-based fusion approaches, its methods naturally handle blends of categorical, continuous, and probabilistic features without extensive explicit normalization or learned embeddings, making it suitable for mixed inputs like weather variables and calibrated vision probabilities~\citep{ke2017_lightgbm,shwartz2022_tabular,grinsztajn2022_trees}. In addition, LightGBM’s bias–variance profile is data-efficient on tabular settings and thus less prone to overfitting than high-capacity deep fusion networks on typical small-to-medium-scale datasets \citep{shwartz2022_tabular,grinsztajn2022_trees}. Further, LightGBM supports feature importance analyses and SHAP-based explanations, enabling attribution of predictive power across visual and meteorological inputs \citep{lundberg2017_shap,lundberg2020_tree_shap}.

\paragraph{YOLO  Visual Feature}
\label{sec:yolo-classifier}
For classification, we fine-tuned the pretrained \texttt{YOLOv8n-cls} model, YOLOv8's smallest and fastest classification variant, on labeled webcam images with visibility categories (\texttt{PERFECT}, \texttt{CLEAR}, \texttt{CLOUDY}, \texttt{OBSCURED}). We excluded images in the \texttt{BAD} class to reduce noise, resulting in 6,149 labeled training examples.

Training used the default YOLOv8 setup: an 80/20 train-validation split, 100 epochs, batch size 16, $224 \times 224$ input resolution. Data augmentation included RandAugment, RandErasing (40\%), Mosaic, and Color Jitter. Optimization was performed with the Adam optimizer (learning rate 0.01, momentum 0.937, weight decay 0.0005).

After finetuning, YOLOv8 was run on the full dataset to produce calibrated class probability outputs for each image. These probability vectors were then used as the vision modality inputs in the late-fusion experiments.

\begin{figure}[ht]
    \centering
    \includegraphics[width=\linewidth]{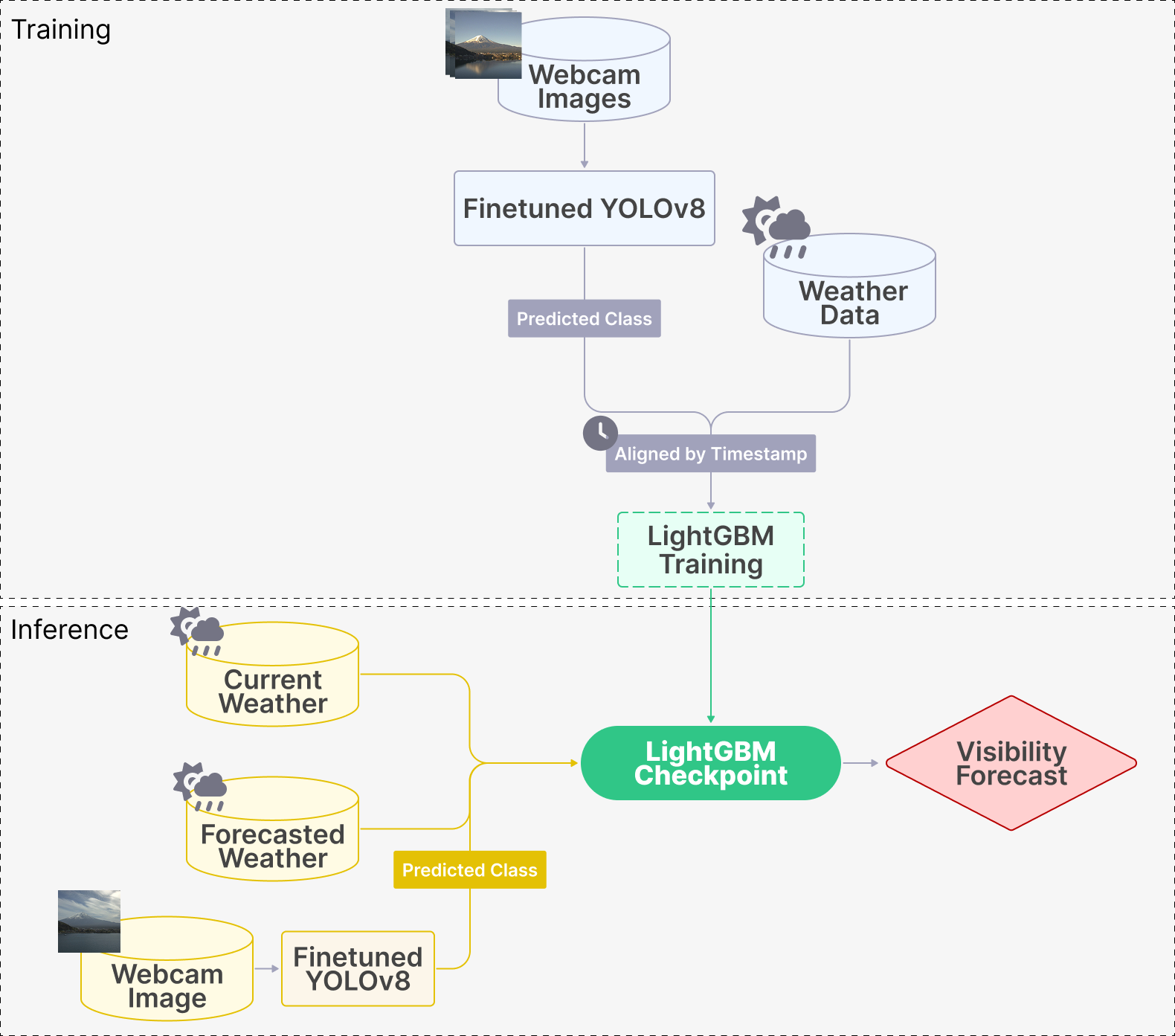}
    \caption{The training and inference process of the late fusion pipeline.}
    \label{fig:lightgbm-late-fusion-pipeline-combined}
\end{figure}
\paragraph{Late-fusion of Features}
\label{sec:lightgbm-inputfeatures}
We organize inputs into three families: visual features (YOLOv8 softmax class probabilities), meteorological features (current and forecasted weather variables in tabular form), and optional metadata features (camera ID, location, time of day). 

For training, we construct a fused dataset (see Figure \ref{fig:lightgbm-late-fusion-pipeline-combined}) where each webcam image is paired with YOLOv8-derived visibility class probabilities (used instead of human labels to prevent data leakage) synchronized by timestamp with contemporaneous and forecasted weather conditions, as well as camera metadata. This multimodal table of features serves as the input to LightGBM. 

At inference time, a single webcam image is passed through YOLOv8 to produce softmax class probabilities, which are fused with current and forecasted weather variables for the camera's location. LightGBM then consumes this fused feature vector to forecast visibility conditions over the same prediction horizons as in training.

We train LightGBM on binary day-level labels indicating whether at least a threshold fraction $\theta$ of that day’s frames are classified as ‘visible’ (CLEAR or PERFECT). In our experiments, we set $\theta = 0.5$, meaning a day is labeled ``visible'' if at least half of its frames meet the criterion. Labels are constructed per (camera, date) and shifted to create horizons (+0d, $\cdots$, +3d). To reduce leakage for +0d, we use either a first-frame snapshot or a 3-hour morning window. We evaluate with 5-fold GroupKFold over (camera, date) using Accuracy (ACC) and ROC-AUC, reporting means across folds.

\section{Results}
\subsection{YOLOv8 Classifier Performance}
\label{sec:results-yolo}
We fine-tuned the pretrained \texttt{YOLOv8n-cls} model on 6,149 labeled webcam images as described in Section~\ref{sec:yolo-classifier}.

\begin{figure}[ht]
    \centering    \includegraphics[width=0.85\linewidth]{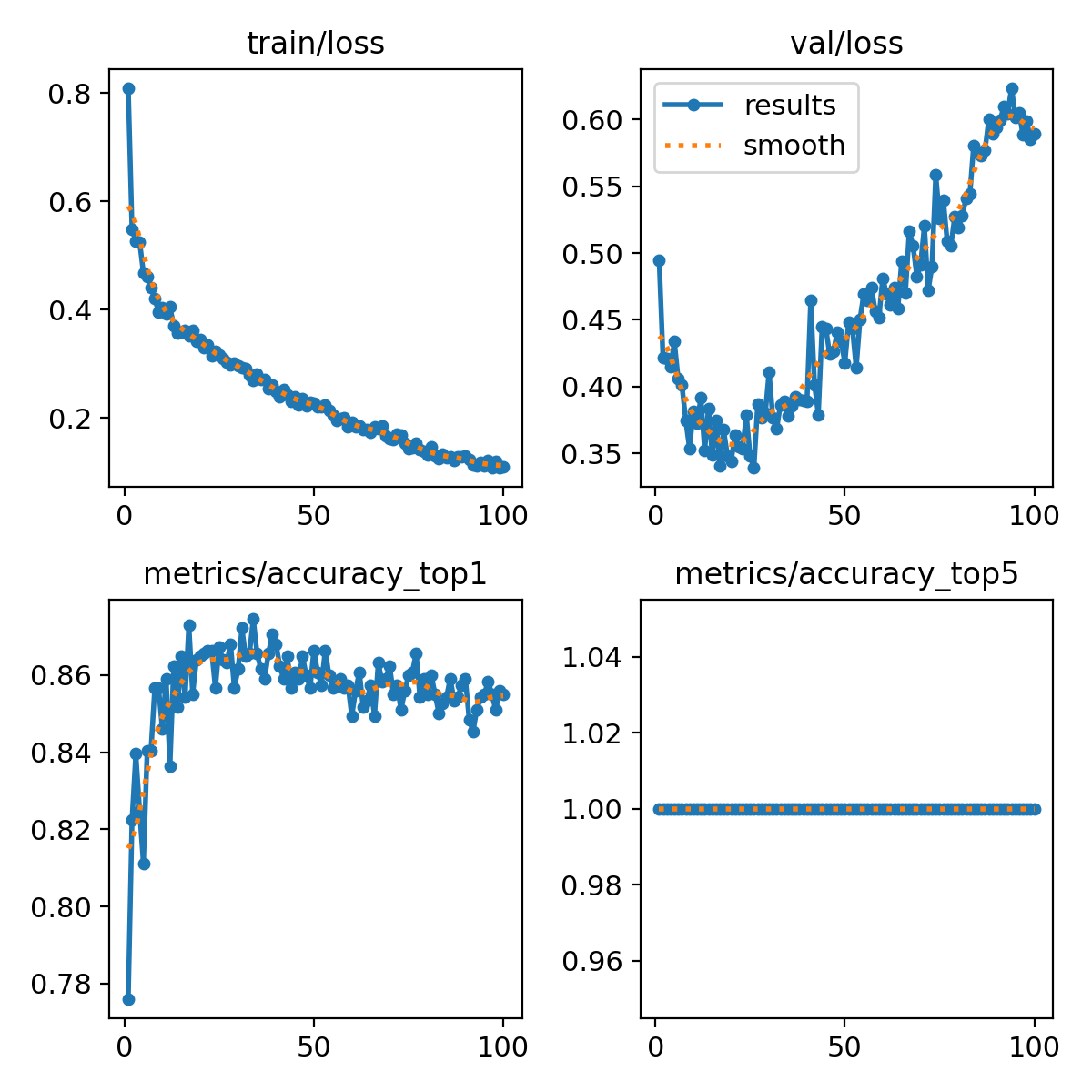}
    \caption{Training dynamics of the YOLOv8n-cls classifier}
    \label{fig:yolo-training}
\end{figure}
The model achieved a peak top-1 validation accuracy of 87.5\% (validation loss 0.389) at epoch 33, after which performance plateaued and validation loss began to rise, consistent with mild overfitting (Figure~\ref{fig:yolo-training}). As expected, top-5 accuracy remained at 1.0 across training due to the limited number of classes.

\begin{figure}
    \centering
    \includegraphics[width=0.85\linewidth]{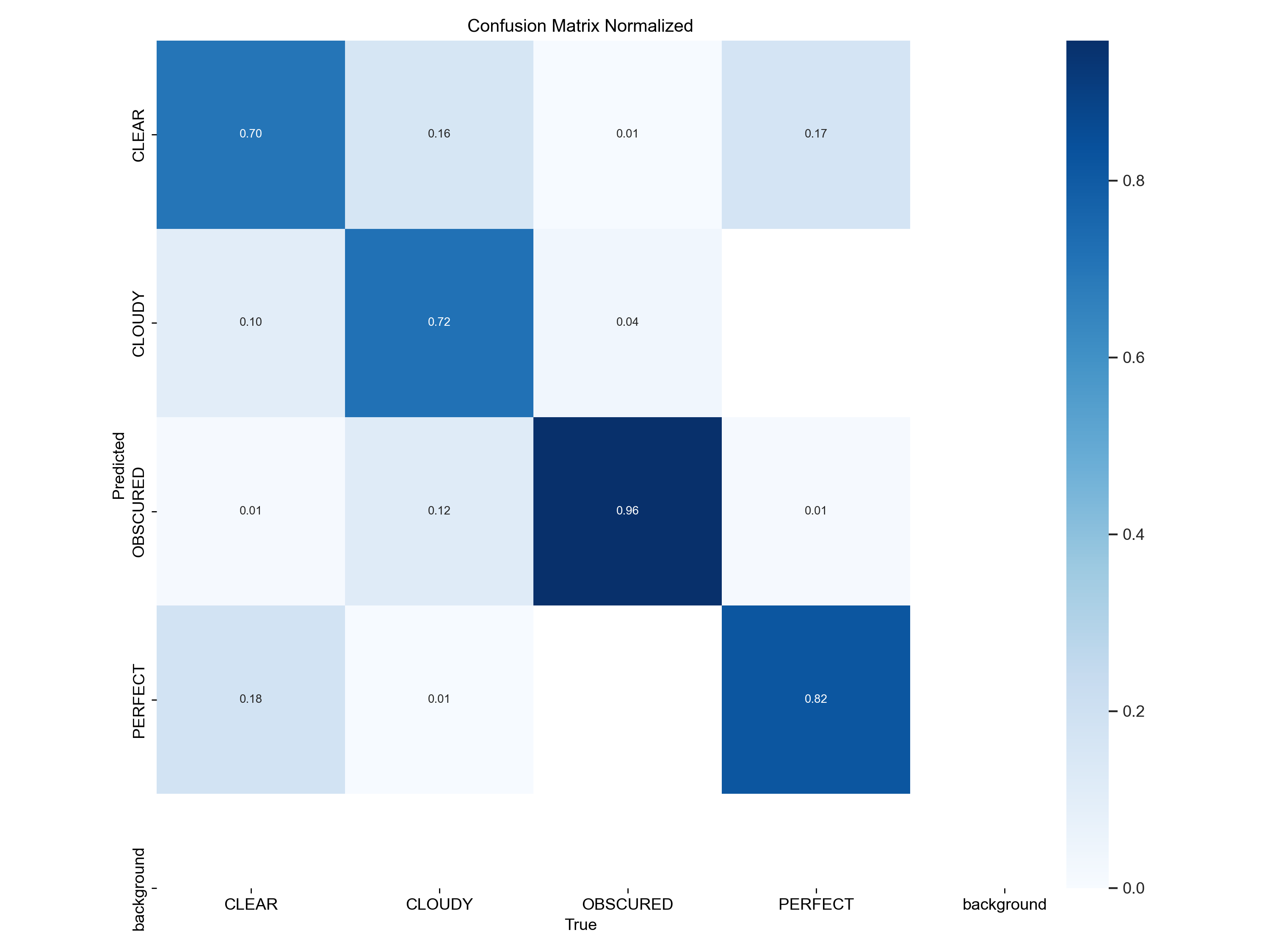}
    \caption{Normalized confusion matrix of YOLOv8n-cls predictions on the validation set.}
    \label{fig:yolo-confusion}
\end{figure}

\begin{figure}
    \centering
    \includegraphics[width=0.8\linewidth]{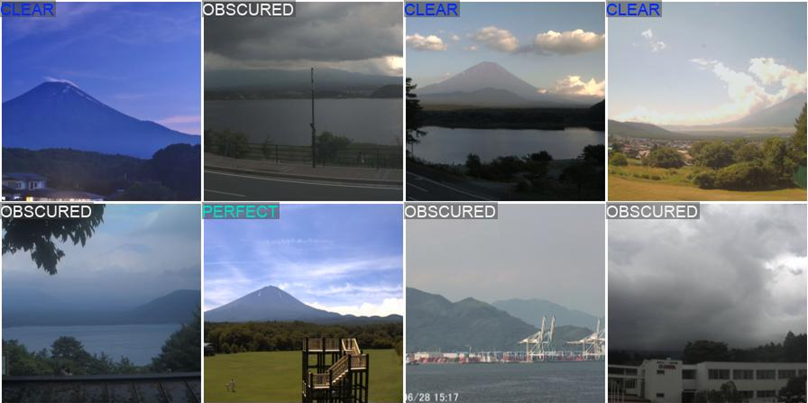}
    \caption{Example YOLOv8 predictions}
    \label{fig:yolo-class-ex}
\end{figure}

To further assess per-class performance, we report the normalized confusion matrix in Figure~\ref{fig:yolo-confusion}. The model demonstrates very high accuracy for the \texttt{OBSCURED} class (96\%), which aligns with the dominance of this class in the dataset and the relative ease of distinguishing severe occlusion. The \texttt{PERFECT} category also achieved strong performance (82\%), with most errors being misclassifications into \texttt{CLEAR}. The most frequent confusions occurred between \texttt{CLEAR} and \texttt{CLOUDY} (70–72\% correct), which is unsurprising given the subjective boundary between “a few clouds” and “mostly cloudy but still visible.” Examples of model classifications generated during YOLO finetuning can be seen in Figure~\ref{fig:yolo-class-ex}.

Overall, these results confirm that YOLOv8 alone provides a highly informative signal for visibility classification, particularly in distinguishing strongly obscured conditions from all other cases. The calibrated softmax probability outputs from this classifier were subsequently used as visual features in the late-fusion experiments.

\subsection{LightGBM Fusion Results}
\label{sec:lightgbm}

To assess the contribution of multimodal fusion, we trained gradient-boosted decision tree classifiers (LightGBM) on three feature configurations: (1) \textbf{YOLO-only}, using calibrated softmax probabilities from our vision model; (2) \textbf{Weather-only}, using current conditions and forecast features up to three days ahead; and (3) \textbf{Late Fusion}, concatenating both modalities. Models were evaluated across four horizons: \emph{samedaycasting} (+0d), \emph{tomorrowcasting} (+1d), and two multi-day forecasts (+2d and +3d). 

\paragraph{Performance Across Horizons.}
Table~\ref{tab:lightgbm-horizon-results} summarizes mean accuracy (ACC)---the percentage of correct predictions---and area under the ROC curve (AUC)---a threshold-independent measure of ranking performance---computed from 5-fold grouped cross-validation. For \textbf{samedaycasting} (+0d), YOLO-only models achieved the highest discrimination (AUC~0.94), indicating that early-morning visually derived features are the most powerful predictor of same-day visibility. Adding weather features produced only marginal gains in accuracy and slightly reduced AUC, suggesting that meteorological variables added uncertainty at this horizon.
\begin{table}[ht]
\centering
\caption{LightGBM performance (5-fold CV) across horizons and feature configurations.
Values are mean accuracy (ACC) and AUC. Bold = best per horizon.}
\label{tab:lightgbm-horizon-results}
\begin{tabular}{c|c|c|c}
\toprule
\textbf{Horizon} & \textbf{Variant} & \textbf{ACC} & \textbf{AUC} \\
\midrule
\multirow{3}{*}{+0d} & YOLO-only       & 0.892 & \textbf{0.940} \\
                     & Weather-only    & 0.747 & 0.741 \\
                     & YOLO+Weather   & \textbf{0.899} & 0.897 \\
\midrule
\multirow{3}{*}{+1d} & YOLO-only       & 0.644 & 0.724 \\
                     & Weather-only    & 0.711 & 0.652 \\
                     & YOLO+Weather   & \textbf{0.741} & \textbf{0.728} \\
\midrule
\multirow{3}{*}{+2d} & YOLO-only       & 0.741 & \textbf{0.707} \\
                     & Weather-only    & 0.766 & 0.500 \\
                     & YOLO+Weather   & \textbf{0.773} & 0.562 \\
\midrule
\multirow{3}{*}{+3d} & YOLO-only       & 0.654 & 0.638 \\
                     & Weather-only    & 0.723 & 0.631 \\
                     & YOLO+Weather   & \textbf{0.715} & \textbf{0.680} \\
\bottomrule
\end{tabular}
\end{table}
At +1d (\textbf{tomorrowcasting}), late fusion was clearly beneficial, increasing accuracy by nearly 10\% over YOLO-only and slightly outperforming weather-only models. This indicates that combining current visual conditions with forecasted variables provides a more robust signal for next-day predictions. 

At +2d, YOLO retained the highest AUC, though fusion improved overall accuracy, suggesting that YOLO features provide a stronger ranking signal but forecasts dominate hard classification. By +3d, fusion offered clear improvements over both single-modality baselines in both accuracy and AUC, as expected given that weather forecasts carry most of the predictive signal at this range.

\paragraph{Effect of Windowing.}
We additionally compared two strategies for constructing daily training examples: a \textbf{first-frame snapshot}, which uses the earliest available frame from each day, and a \textbf{3-hour morning window}, which averages YOLO probabilities and weather features across all frames captured between midnight and 03:00. The windowed approach aims to denoise early-morning noise (e.g., spurious frames affected by darkness or transients) by aggregating multiple observations into a single, more stable feature vector.

Table~\ref{tab:window-effect} reports the difference in performance between these strategies ($\Delta = \text{Window} - \text{First}$). Windowing substantially improved YOLO-only discrimination at both $+0$d and $+3$d ($\Delta \mathrm{AUC} \approx +0.07$), indicating that multiple-frame aggregation helps YOLO better rank same-day and longer-horizon visibility. However, fusion models experienced modest drops in accuracy at $+2$d ($\Delta \mathrm{ACC} \approx -0.08$) and $+3$d ($\Delta \mathrm{ACC} \approx -0.15$), suggesting that averaging may oversmooth informative outlier frames that help longer-horizon forecasts. These results point to a trade-off between temporal stability and early availability: windowing is highly beneficial for pure vision models, but the optimal fusion strategy may require retaining frame-level extremes or using rolling updates later in the day.

\begin{table}[ht]
\centering
\caption{Effect of using only the first daily frame vs. aggregating frames within a 3-hour morning window. All results are based on the late fusion of features.}
\label{tab:window-effect}
\begin{tabular}{c|c|c|c|c}
\toprule
\multirow{2}{*}{\textbf{Horizon}} & \multicolumn{2}{c|}{\textbf{ACC}} & \multicolumn{2}{c}{\textbf{AUC}} \\ 
\cmidrule(lr){2-3} \cmidrule(lr){4-5}
 & \textbf{First} & \textbf{Window} & \textbf{First} & \textbf{Window} \\ 
\midrule
\textbf{+1d} & \textbf{0.84} & 0.74 & 0.73 & \textbf{0.73} \\
\textbf{+2d} & \textbf{0.77} & 0.77 & \textbf{0.56} & 0.56 \\
\textbf{+3d} & \textbf{0.72} & 0.72 & 0.68 & \textbf{0.68} \\
\bottomrule
\end{tabular}
\end{table}

\begin{figure}
    \centering
    \includegraphics[width=\linewidth]{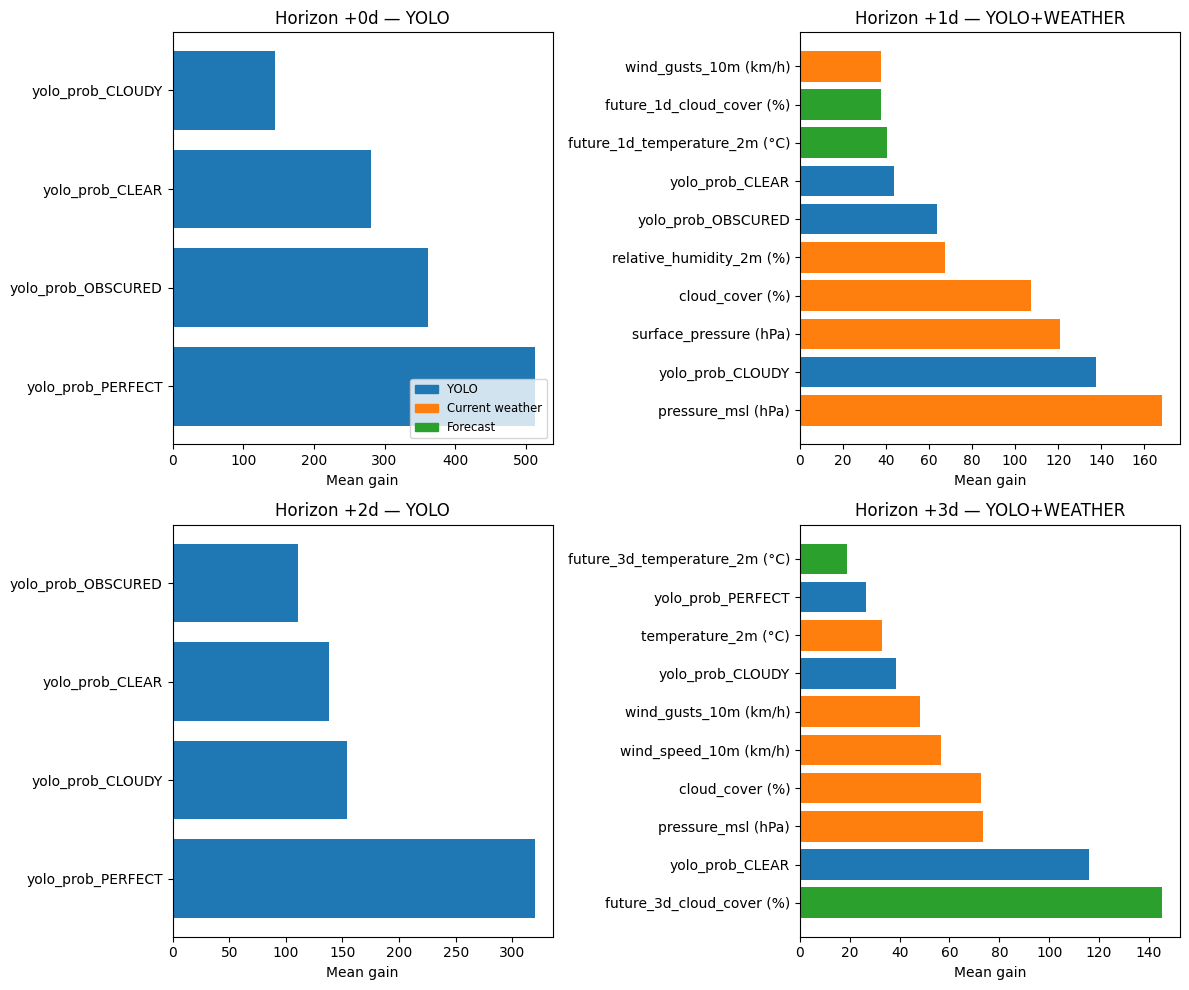}
    \caption{Top features (mean gain) for the best-performing variant at each horizon. 
  Bars are colored by modality: YOLO (blue), current weather (orange), and forecast features (green).
  The model shifts from vision-dominated at +0d to forecast-dominated by +3d, with fusion mediating the mid-range horizons.}
    \label{fig:fi-horizons}
\end{figure}

\paragraph{Feature Importance.}
Feature importance analysis (Figure~\ref{fig:fi-horizons}) confirms these trends. For +0d, YOLO probabilities—especially $P(\texttt{PERFECT})$ and $P(\texttt{OBSCURED})$—dominate the model, with surface pressure providing secondary information. For +1d and beyond, forecasted cloud cover becomes the single most important feature, followed by pressure and humidity variables. Importantly, YOLO probabilities remain non-negligible contributors even at +1d and +2d, indicating that the visual state of the mountain at the time of prediction adds information not fully captured by numerical forecasts. In other words, webcam imagery serves as a "reality check" for the forecast, allowing the model to adjust predictions when actual conditions deviate from what the forecast suggests (e.g., if Fuji is already obscured earlier than expected).

Overall, these results demonstrate that YOLO-based vision features are indispensable for short-term forecasts, whereas weather variables become increasingly important beyond one day. Late fusion achieves the most balanced performance at horizons $\geq$1 day and should be preferred for practical deployment.

\subsection{Insights into dataset}
\label{sec:dataset-insights}
Our experiments provided several insights into the characteristics and challenges of the FujiView dataset. First, class imbalance remains a central issue with \texttt{OBSCURED} conditions dominating, reflecting the reality that Mount Fuji is often hidden by cloud cover, particularly during the rainy season when the current dataset was collected. This imbalance may bias models toward predicting non-visibility.

Second, we observed that horizon difficulty is non-monotonic: tomorrowcasting (+1d) was often easier than todaycasting (+0d), despite having a longer prediction window. This may reflect the stability of synoptic-scale weather patterns captured by forecasts, whereas same-day visibility is more sensitive to transient cloud dynamics not fully captured in input features.

Finally, the comparison between first-frame and windowed snapshots highlights that using a short early-morning window can denoise YOLO predictions and boost AUC for +0d and +3d tasks, but may slightly reduce accuracy for +2d and +3d fusion models. This suggests a tradeoff between temporal coverage and early availability: while windowing stabilizes vision features, it may also smooth out informative early-morning extremes that help longer-horizon forecasts. Future work could explore incremental or rolling prediction updates throughout the day, and continued data collection will help clarify whether these windowing effects persist as the dataset becomes more complete.

\section{Discussion}
\label{sec:discussion}

\paragraph{Novelty and Positioning.}
Prior work on visibility estimation has either focused on single-modality vision models applied to webcam streams, or on numerical weather prediction models that ignore perceptual visibility altogether. FujiView is the first system to fuse both modalities at scale for the explicit task of scenic visibility. In doing so, it introduces three distinct forms of novelty: (1) the definition of \emph{Scenic Visibility Forecasting (SVF)} as a benchmark task centered on human-perceived visibility outcomes rather than raw meteorological variables; (2) a continually expanding multimodal dataset, with over 100,000 images already collected, ~26,000 human-labeled instances to date, and more than 320,000 images projected by year's end; and (3) a principled late-fusion framework that demonstrates strong baseline performance and interpretability across different forecasting horizons. By positioning SVF as a benchmark task, FujiView provides both a concrete dataset+pipeline and a conceptual foundation for future multimodal research that extends beyond Fuji to other landmarks and forecasting setups.

The FujiView dataset and modeling framework provide a platform for broader environmental forecasting tasks where human-centric visibility is important (e.g., air quality monitoring, wildfire smoke impacts, or landmark-based navigation). Because the dataset is continually expanding, FujiView is positioned not only as a tool for end users but also as a living benchmark that researchers and practitioners can build upon, accelerating progress at the intersection of multimodal learning and environmental decision support.

\paragraph{Limitations.}
The current FujiView dataset was collected between April and August, during Japan's rainy season. Unsurprisingly, this period skews heavily toward the \texttt{OBSCURED} class (see Figure \ref{fig:class-dist-pie}), reflecting the reality that Mount Fuji is rarely clear in these months. However, even with year-round collection, some imbalance is inevitable since Fuji is obscured more often than not. Such imbalance may bias calibration and complicate minority class performance.

A second limitation arises from camera heterogenity and seasonal drift. Webcams differ in resolution, angle, exposure settings, and maintenance, while season shifts (snow cover, vegetation, daylight hours, etc) alter the visual domain. Both factors challenge model generalization across camera sites and time. 

Third, forecasting inherits the uncertainty of numerical weather prediction (NWP). Errors in NWP propagate directly into visibility forecasts, limiting the ceiling on predictive accuracy regardless of the fusion model.

Finally, nighttime handling remains problematic. Nearly all webcams saturate or underexpose low-light conditions, rendering Mount Fuji not visible independent of actual weather. Consequently, we excluded full-darkness frames, but this decision sacrifices potentially informative nocturnal patterns (e.g., evening cloud buildup, pre-dawn clearing) that could improve forward prediction. Developing methods that can leverage nighttime conditions remains an open challenge, one that likely requires direct control over the deployment and configuration of new camera sites.


\paragraph{Future Works.}
Within this benchmark space, temporal aggregation methods (e.g., ConvLSTMs or Temporal Fusion Transfomers) could leverage all observations up to a cutoff time to better capture weather inertia, while counterfactual features such as forecast deltas and persistence patterns may refine longer-horizon predictions. Beyond Fuji, extending SVF to other landmarks (e.g., Matterhorn, Rainier) would establish a cross-site multimodal benchmark. We also see opportunities in self-supervised vision pretraining on continuous webcam streams, and in uncertainty-aware decision layers (e.g., selective prediction) that better reflect real-world traveler needs.

\section{Conclusion}

FujiView demonstrates that simple, principled late fusion of webcam imagery and structured meteorological data provides robust visibility forecasting across horizons from same-day (+0d) to three days ahead (+3d). Our results show that YOLO-based vision features dominate short-term predictions and remain valuable even at longer horizons, while weather-driven forecasts increasingly take over as the primary signal beyond +1d. Late fusion consistently matches or outperforms either modality alone, particularly improving accuracy and AUC for +1d and +3d forecasts. Finally, our comparison of first-frame versus windowed snapshots suggests that a short early-morning window improves ranking ability for samedaycasts without degrading longer-horizon performance, making it a practical choice for real-time deployment.

Beyond these experimental findings, this work introduces \emph{Scenic Visibility Forecasting (SVF)} as a new benchmark task for multimodal learning with opportunities to incorporate temporal modeling, richer input horizons, and cross-site generalization to other landmarks. By formalizing visibility prediction as a human-centered forecasting challenge, and by releasing a continually expanding dataset of over 100,000 (projected 300,000+) Fuji webcam images paired with aligned forecasts, we establish a foundation for reproducible and extensible research.

{
    \small
    \bibliographystyle{ieeenat_fullname}
    \bibliography{main}
}

\end{document}